\documentclass[runningheads]{llncs}
\usepackage{graphicx}
\usepackage{times}
\usepackage{epsfig}
\usepackage{amsmath}
\usepackage{amssymb}
\usepackage{xspace}

\usepackage[inline,shortlabels]{enumitem}
\usepackage{soul}
\usepackage{subfigure}
\usepackage{verbatim}
\usepackage[pagebackref=true,breaklinks=true,letterpaper=true,colorlinks,bookmarks=false]{hyperref}
\usepackage{changepage}
\usepackage{booktabs}
\usepackage{multirow}
\usepackage{wrapfig}
\makeatletter
\DeclareRobustCommand\onedot{\futurelet\@let@token\@onedot}
\def\@onedot{\ifx\@let@token.\else.\null\fi\xspace}

\def\eg{\emph{e.g}\onedot}

\def\etal{\emph{et al}\onedot}
\makeatother

\begin{document}
\title{Semi-supervised Learning Approach to Generate Neuroimaging Modalities with Adversarial Training}
\titlerunning{SSL to Generate Neuroimaging Modalities with Adversarial Training}
%

\author{Harrison Nguyen\inst{1}\and
Simon Luo\inst{1}\inst{2} \and
Fabio Ramos\inst{1}}
\authorrunning{H. Nguyen et al.}
%
\institute{The University of Sydney, The University of Sydney NSW 2006, Australia \email{\{harrison.nguyen,fabio.ramos\}@sydney.edu.au}\and
\email{sluo4225@uni.sydney.edu.au}\\}

\maketitle              
\begin{abstract}
Magnetic Resonance Imaging (MRI) of the brain can come in the form of different modalities such as T1-weighted and Fluid Attenuated Inversion Recovery (FLAIR) which has been used to investigate a wide range of neurological disorders. Current state-of-the-art models for brain tissue segmentation and disease classification require multiple modalities for training and inference. However, the acquisition of all of these modalities are expensive, time-consuming, inconvenient and the required modalities are often not available. As a result, these datasets contain large amounts of \emph{unpaired} data, where examples in the dataset do not contain all modalities. On the other hand, there is smaller fraction of examples that contain all modalities (\emph{paired} data) and furthermore each modality is high dimensional when compared to number of datapoints. In this work, we develop a method to address these issues with semi-supervised learning in translating between two neuroimaging modalities. Our proposed model, Semi-Supervised Adversarial CycleGAN (SSA-CGAN), uses an adversarial loss to learn from \emph{unpaired} data points, cycle loss to enforce consistent reconstructions of the mappings and another adversarial loss to take advantage of \emph{paired} data points. Our experiments demonstrate that our proposed framework produces an improvement in reconstruction error and reduced variance for the pairwise translation of multiple modalities and is more robust to thermal noise when compared to existing methods.

\end{abstract}

\section{Introduction}
\begin{figure}[t]
\centering
\scalebox{0.73}{
   \hspace{-0.65cm}\includegraphics[width=\linewidth]{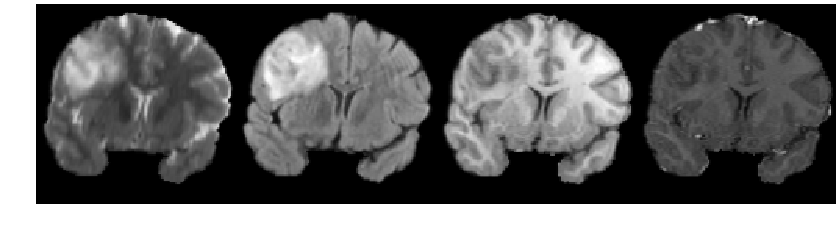}
   }
\scalebox{0.7}{
    \includegraphics[width=\linewidth]{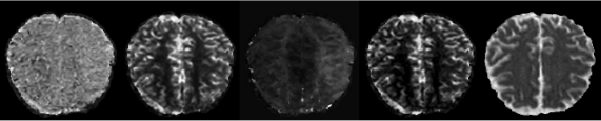}
 }
   \caption{\textbf{Top}: A coronal slice of a low grade glioma (brain tumour) in the BraTs dataset in different modalities. From left to right: T2, Fluid-attenuated inversion recovery (FLAIR), T1 and T1c. \textbf{Bottom}: Axial slices of modalities of a CT perfusion scan of an ischemic stroke lesion patient in the ISLES dataset. From left to right: Mean Transit Time (MTT),cerebral blood flow (CBF),time to peak of the residue function (Tmax), cerebral blood volume (CBV),Apparent diffusion coefficient (ADC).}
\label{fig:long}
\end{figure}

Magnetic Resonance Imaging (MRI) of the brain has been used to investigate a wide range of neurological disorders and depending on the imaging sequence used, can produce different modalities such as T1-weighted images, T2-weighted images, Fluid Attenuated Inversion Recovery (FLAIR), and diffusion weighted imaging (DWI). Each of these modalities produce different contrast and brightness of brain tissue that could reveal pathological abnormalities. Many of the advances in the use of data-driven models in Alzheimer's disease classification\cite{lu2018multimodal}, brain tumour segmentation\cite{havaei2017brain}  and skull stripping methods\cite{mahbod2018automatic}, rely on deep convolutional neural networks (DCNN). In particular, datasets such as BraTs\cite{menze2015multimodal} and ISLES\cite{maier2017isles} have been focusing on the evaluation of state-of-the-art methods for the segmentation of brain tumours and stroke lesions respectively. These methods do not require the use of hand designed features and instead are able to learn a hierarchy of increasingly complex features. However, they require multiple neuroimaging modalities for high performance and improved sensitivity\cite{dai2012classification} (See Figure \ref{fig:long}). Collecting multiple modalities for each patient can be difficult, expensive and not all of these modalities are available in clinical settings. In particular, \emph{paired} data, where an example has all modalities present, is difficult to access, making these data dependent models more difficult to train or reduce their applicability during inference.

To ensure each modality is present, the missing modality could be imputed through a domain adaptation model where characteristics of one image set is transferred into another image set (\eg T1-weighted to T2-weighted) that has been learned from existing \emph{paired} examples. However, since this \emph{paired} data is limited in the neuroimaging context, learning from examples that do not have all modalities (\emph{unpaired} data) is valuable as this form of data is more readily available.

There has been significant interest in unsupervised image-to-image translation where \emph{paired} training data is not available but two distinct image sets. Methods proposed by Zhu \etal \cite{zhu2017unpaired} and Hoffman \etal \cite{hoffman2017cycada} assume the two image collections are representations of some shared, underlying state. They use adversarial training which discriminates at the image level to guide the transformation between the domains. Furthermore, the translations between these two sets should have approximately invertible solutions and should be \textit{cycle consistent}- where the mapping of a particular source domain to the target domain and back should yield the original source at the pixel level. Alternative methods extract domain invariant features with DCNNs and discriminate the feature distributions of source/target domains \cite{tzeng2017adversarial}.

One work in recent literature that exploits the two distinct image sets of \emph{unpaired} data, in order to improve the performance on tasks with a scarcity of \emph{paired} data is the Cycle Wasserstein Regression GAN\cite{mcdermott2018semi} (CWRG). The CWRG uses the $l2$-norm as a penalty term for the reconstruction of \emph{paired} data along with the adversarial signal and cycle-loss of the CycleGAN. However, the CWRG demonstrated its performance on ICU timeseries data and transcriptomics data and not on image data. 

Our proposed method, the \emph{Semi-Supervised Adversarial CycleGAN} (\emph{SSA-CGAN}) further extends the application of leveraging \emph{unpaired} data and \emph{paired} data to MRI image translation, where the dimensionality of the examples is orders of magnitude larger. Our method uses multiple adversarial signals for semi-supervised bi-directional image translation. Our experimental results have demonstrated that our proposed approach has superior performance compared to the CycleGAN and CWRG in terms of average reconstruction error and variance and as well as robustness to noise when evaluated using the BraTs and ISLES dataset.

\section{Related Work}
General adversarial networks (GAN) have received significant attention since the work by \cite{goodfellow2014generative} and various GAN-based models have achieved impressive results in image generation \cite{denton2015deep} and representation learning \cite{radford2015unsupervised}. These models learn a generator to capture the distribution of real data by introducing a competing model, the discriminator, that evolves to distinguish between the real data and the fake data produced by the generator. This forces the generated image to be in distinguishable from real images.

Various conditional GANs (cGAN) have been adapted to condition the image generator on images instead of a noise vector to be used in applications such as  style transfer from normal maps to images \cite{wang2016generative}. Isola \etal's\cite{isola2017image} work in particular, uses labeled image pairs to train a cGAN to learn a mapping between the two image domains. On the other hand, there have been significant works that have tackled image-to-image translation in the unpaired setting. The CycleGAN \cite{zhu2017unpaired} uses a \emph{cycle consistency loss} to ensure the forward mapping and back results in the original image. It has demonstrated success in tasks where paired training data is limited \eg in painting style and season transfer. The Dual GAN, being inspired by dual learning in machine translation used a similar loss objective, where the reconstruction error is used to measure the disparity between the reconstructed object and the original \cite{yi2017dualgan}. Unlike the previous two frameworks, the CoGAN\cite{liu2016coupled} and cross-modal scene networks\cite{aytar2018cross} does not use a cycle consistency loss but instead, uses weight sharing between the two GANs, corresponding to high level semantics to learn a common representation across domains. 

GANs have been used in the semi-supervised learning (SSL) context as the visually realistic images generated can be used as additional training data. Salimans \etal \cite{salimans2016improved} proposed techniques to improve training GANs which included learning a discriminator on additional class labels which can be used for SSL. Mayato \etal \cite{miyato2018virtual} modified the adversarial objective to a regularization method based on virtual adversarial loss. The method probabilistically produces labels that are unknown to the user and computes the adversarial direction based on the virtual labels. Park \etal \cite{park2018adversarial} improves upon the performance of virtual adversarial training by using adversarial dropout which  maximizes the divergence between the training supervision and the outputs from the network with the dropout.

GANs have been used in a range of applications in biomedical imaging such as the generation of multi modal MRI images and retinal fundus images \cite{beers2018high}, to detect anomalies in retinal OCT images\cite{seebock2018unsupervised} and image synthesis of MR and CT images\cite{yang2018unpaired}. Adversarial methods have also been extended to domain adaptation for medical imaging. Chen \etal \cite{chen2019synergistic} recently developed the Synergistic Image and Feature Adaptation framework that enhances domain-invariance through feature encoder layers that are shared by the target and source domain and uses additional discriminator to differentiate the feature distributions. Perone \etal forgoes the use of adversarial training and instead demonstrates application of self ensembling and mean teacher framework \cite{perone2018unsupervised}. 

The CycleGAN has been recently applied to the biomedicial field for translating between sets of data. Welander \etal \cite{welander2018generative} investigated the difference between the CycleGan and UNIT\cite{liu2017unsupervised} for the translation between T1 and T2 MRI modalities and found the CycleGAN was the better alternative if the aim was to generate visually realistic images as possible. McDermott \etal \cite{mcdermott2018semi} on the other hand, tackled domain adaptation in the semi-supervised setting by proposing Wasserstein CycleGANs coupled with a $l_2$ regression loss function on paired data. The semi-supervised setting for this paper is similar to McDermott \etal, however we propose an adversarial training signal for paired data instead of the $l_2$ loss. We demonstrate our method produces better reconstructions with lower variance and is more robust to noise in the context of translating between neuroimaging modalities compared to existing methods.
\section{Methods}

\begin{figure*}
\centering
\vspace{-1cm}
\includegraphics[width=0.7\linewidth,height=4cm]{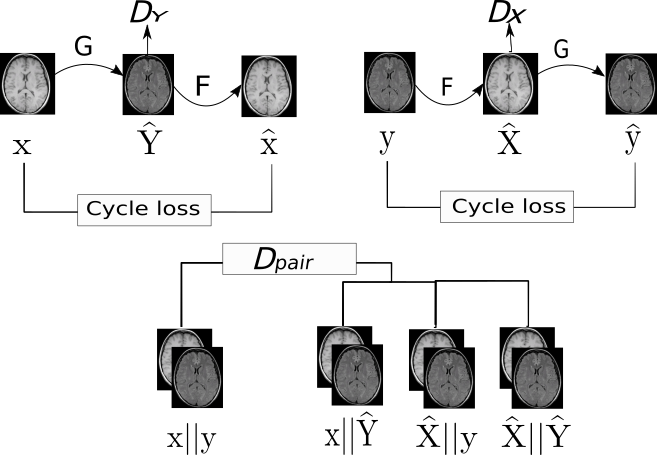}
   \caption{Our model is composed of the CycleGAN architecture and an axuillary discriminator which takes as input concatenated paired examples and the concatenation of generators' various transformations.}
\label{fig:short}
\end{figure*}
\vspace{-1cm}
\subsection{CycleGAN}
The CycleGAN\cite{zhu2017unpaired} learns to translate points between two domains $X$ and $Y$. Given two sets of unlabeled and \emph{unpaired} images,$\{ x_i\}^N_{i=1}$ where $x_i \in X$ and $\{ y_j\}^M_{j=1}$, $y_j \in Y$, two generators, $F$ and $G$, are trained to learn mapping functions $G: X\rightarrow Y$ and $F:Y\rightarrow X$, where $F$ and $G$ are usually represented by DCNNs. Furthermore, two discriminators $D_X$ and $D_Y$ are trained where $D_X$ learns to distinguish between images $\{x\}$ and $\{F(y)\}$ and $D_Y$ discriminates between $\{y\}$ and $\{G(x)\}$. Instead of the original GAN loss, the CycleGAN trains discriminators using the least squares loss function proposed by Mao \etal\cite{mao2017least}. For example, $D_X$ minimises the following objective function:
\begin{equation}
\begin{split}
    \mathcal{L}_{D_{X}} &= \mathbb{E}_{\mathbf{x}\sim P(\mathbf{x})}\big[(D_X(\mathbf{x}) - 1)^2\big] 
    + \mathbb{E}_{\mathbf{y}\sim P(\mathbf{y})}\big[(D_X(F(\mathbf{y})))^2\big].
\end{split}
\end{equation}

Conversely the generator, $F$, for example is trained according to the following \emph{adversarial loss},

\begin{equation}
     \mathcal{L}_{F_{adv}} = \mathbb{E}_{\mathbf{y}\sim P(\mathbf{y})}\big[(D_X(F(\mathbf{y}))-1)^2\big],
\end{equation}
as well as a \emph{cycle-consistency loss} where reconstruction error between the inverse mapping and the original point is minimised\cite{zhu2017unpaired},
\begin{equation}
\begin{split}
\mathcal{L}_{cyc} &= \mathbb{E}_{\mathbf{x}\sim P(\mathbf{x})}\big[||F(G(\mathbf{x})) - \mathbf{x}||_1] + \mathbb{E}_{\mathbf{y}\sim P(\mathbf{y})}\big[||G(F(\mathbf{y})) - \mathbf{y}||_1].
\end{split}
\end{equation}
The overall loss function for the generator is therefore given as
\begin{equation}
    \mathcal{L}_{F} =  \mathcal{L}_{F_{adv}} + \lambda\mathcal{L}_{cyc},
\end{equation}
where $\lambda$ controls the relative strength between the adversarial signal and the cycle-consistency loss.

\subsection{Semi-Supervised Adversarial CycleGAN}
We extend the CycleGAN through the Semi-Supervised Adversarial CycleGAN (SSA-CGAN) to take advantage of \emph{paired} training data. In our scenario we have additional information in the form of $T$ paired examples $\{\mathbf{x}_p,\mathbf{y}_p\}^T_{p=1}$, a subset $P \subseteq X \times Y$. We seek to take advantage of this paired information through an auxiliary adversarial network, $D_{pair}$ (See Figure \ref{fig:short}). $D_{pair}$ takes as input, only the paired examples from $P$ and the concatenations of the following transformations:
\begin{enumerate*}[a)]
  \item $\mathbf{x}_p$ and $\mathbf{y}_p$,
  \item $\mathbf{x}_p$  and  $G(\mathbf{x}_p)$,
  \item $F(\mathbf{y}_p)$ and  $\mathbf{y}_p$,
  \item $F(\mathbf{y}_p)$ and  $G(\mathbf{x}_p)$.
\end{enumerate*} 
$D_{pair}$ attempts to discriminate between the ground-truth pairs, $\{\mathbf{x}_p,\mathbf{y}_p\}\in P$, as real and the transformation of the image and its respective real image as fake. 
Therefore, the paired discriminator minimises
\begin{equation}
    \begin{split}
            \mathcal{L}_{D_{pair}}&= \mathbb{E}_{\mathbf{x},\mathbf{y}\sim P_{pair}(\mathbf{x},\mathbf{y})}\big[(D_{pair}(\mathbf{x},\mathbf{y})-1)^2\big]+\frac{1}{3}\Big[\mathbb{E}_{\mathbf{x},\mathbf{y}\sim P_{pair}}\big[ D_P(\mathbf{x},G(\mathbf{x}))^2\big]+\\
              &\mathbb{E}_{\mathbf{x},\mathbf{y}\sim P_{pair}}\big[D_{pair}(F(\mathbf{y}),\mathbf{y})^2\big]+
            \mathbb{E}_{\mathbf{x},\mathbf{y}\sim P_{pair}}\big[D_{pair}(F(\mathbf{y}),G(\mathbf{x}))^2\big]\Big] \\
    \end{split}
\end{equation}

and $F$'s loss is
\begin{equation}\label{eqn:full_objective}
        \mathcal{L}_{F_{Semi}} = \mathcal{L}_{F_{adv}} + \lambda\mathcal{L}_{cyc}
            + \alpha \mathcal{L}_{pair},
\end{equation}
where $\mathcal{L}_{pair}$ is given as
\begin{equation}
    \begin{split}
    \mathcal{L}_{pair}&=\mathbb{E}_{\mathbf{x},\mathbf{y}\sim P_{pair}}\big[ (D_{pair}(\mathbf{x},G(\mathbf{x}))-1)^2\big]+\mathbb{E}_{\mathbf{x},\mathbf{y}\sim P_{pair}}\big[(D_{pair}(F(\mathbf{y}),\mathbf{y})-1)^2\big] \\
            &+\mathbb{E}_{\mathbf{x},\mathbf{y}\sim P_{pair}}\big[(D_{pair}(F(\mathbf{y}),G(\mathbf{x}))-1)^2\big]. \\
    \end{split}
\end{equation}

and $\alpha$ and $\lambda$ control the relative weight of the losses.
The third loss term can be seen as further regularisation of the generators where its forward and backward transformations are pushed towards the joint distribution of $X$ and $Y$.

\section{Experiments}
\subsection{Dataset}
We evaluate our method using BraTS and ISLES datasets which have been used to evaluate state-of-the-art methods for the segmentation of brain tumours and lesions respectively.
BraTS utilizes multi-institutional pre-operative MRI scans and focuses on the segmentation of intrinsically heterogeneous (in appearance, shape, and histology) brain tumors, namely gliomas. This proposed method is trained and tested on the BraTs 2018 dataset. The training dataset contains 285 examples including 210 High GradeGlioma (HGG) cases and 75 cases with Low Grade Glioma (LGG). For each case, there are four MRI sequences, including the T1-weighted (T1), T1 with gadolinium enhancing contrast (T1c), T2-weighted (T2) and FLAIR. The dataset includes pre-processing methods such as skullstrip, co-register to a common space and resample to isotropic $1mm\times1mm\times1mm$ resolution. Bias field correction is done on the MR data to correct the intensity in-homogeneity in each channel using N4ITK tool\cite{tustison2010n4itk}. 

The dataset was divided as the following: 30\% of examples was designated as \emph{unpaired} examples of domain $X$ (\eg T2-weighted volumes) and 30\% as \emph{unpaired} examples of domain $Y$ (\eg T1-weighted), 10\% was designated as \emph{paired} training examples where each example, for example, had both T2-weighted and T1-weighted modalities. 10\% was reserved as a held-out validation set for hyperparameter tuning and 20\% was reserved to be a test set used for evaluation.  

ISLES contains patients who have received the diagnosis of ischemic stroke by MRI. Ischemic stroke is the most common cerebrovascular disease and one of the most common causes of death and disability worldwide\cite{world2012cause}. The stroke MRI was performed on either a 1.5T (Siemens Magnetom Avanto) or 3T MRI system (Siemens Magnetom Trio). Sequences and derived maps were cerebral blood flow (CBF), cerebral blood volume (CBV), time-to-peak (TTP), and time-to-max (Tmax) and mean transit time (MTT). The dataset included images that were rigidly registered to the T1c with constant resolution of $2mm\times2mm\times2mm$ and automatically skull-stripped\cite{maier2017isles}. The dataset includes 38 patients in total and was divided in similar proportions as the BraTS experiment regime. 

Further pre-processing for each dataset included each image modality was normalized by subtracting the mean and dividing by the standard deviation of the intensities within the volume and rescaled to values between $1$ and $-1$. The volumes were reshaped to $240\times240$ coronal and $128\times128$ axial slices for the BraTS and ISLES dataset respectively. This resulted in an average of 170 slices per patient for the BraTS dataset and 18 slices per patient in ISLES. 

\subsection{Implementation}\label{implementation}
\textbf{Network Architecture}: The generator network was adapted from Johnson \etal
\cite{johnson2016perceptual} and Zhu \etal\cite{zhu2017unpaired}. The network contains two stride-2 convolutions, 6 residual blocks\cite{he2016deep} and two fractionally strided convolutions with stride $\frac{1}{2}$. The single input discriminator networks is a PatchGAN. The paired input discriminator was a two stride-2 convolution layers. It used the concatenation of feature maps from the second last layer of $D_X$ and $D_Y$ as inputs as a form of weight sharing with the single image discriminators.

\textbf{Training details}: For all the experiments, we set $\lambda =10$ and $\alpha =2$ in Equation \ref{eqn:full_objective} chosen by the performance on the held out validation set averaged across the pairs of MR modalities mentioned in Section \ref{Evaluation_metrics}.  All networks were trained from scratch using NVIDIA V100 GPU with an initial learning rate of $2\times10^{-4}$, weights were initialised using Glorot initialization\cite{glorot2010understanding} and optimised using Adam\cite{kingma2014adam} with a batch size of 1. The learning rate was kept constant for the first 100 epochs and was linearly decreased thereafter to a learning rate of $2\times10^{-7}$. Training was finished after 200 epochs. While standard data augmentation procedures randomly shift, rotate and scale images, the images were only augmented by random shifting during training as the volumes were normalised to the same orientation and shape due to co-registration. 

\subsection{Evaluation metrics}\label{Evaluation_metrics}
We evaluated the \emph{SSA-CGAN} by learning a separate model for the following pairs of MR modalities: T2$\rightarrow$T1, T2$\rightarrow$T1c, T2$\rightarrow$FLAIR, CBF$\rightarrow$MTT, CBF$\rightarrow$CBV, CBF$\rightarrow$TTP, CBF$\rightarrow$Tmax. For example, T2$\rightarrow$T1 indicates the models were evaluated on the reconstruction of a T1 volume when transformed from a T2 volume. This was evaluated against the CycleGAN and the Cycle Wasserstein Regression GAN\cite{mcdermott2018semi} (CWRG) which is currently the only other method in recent literature that combines \emph{unpaired} and \emph{paired} training data for translation between different modalities. We also included in our experiments using the \emph{SSA-CGAN} framework using only \emph{paired} data, labelled SSA-CGAN-p. On the other hand, our proposed method, \emph{SSA-CGAN} uses \emph{paired} data and leverages \emph{unpaired} data to improve learning. The hyperparameter settings for each method is similar to the training details mentioned in Section \ref{implementation}. For each transformation (\eg T2$\rightarrow$T1c) and for each method, five networks were learned, each with different initialization of weights. These models were compared based on two quantitative metrics, the  mean squared error (MSE) and mean absolute error (MAE) averaged across the five runs and its standard deviation. 

\subsection{Results}

\begin{table}
\begin{center}
\begin{adjustwidth}{-2.0cm}{}
\scalebox{0.9}{
\begin{tabular}{c| c||c|c|c|c|c|c|c}
\toprule
 &Method & T1 &T1c & FLAIR &MTT& rCBV &  TTP & Tmax\\
\midrule
 \multirow{3}{*}{\rotatebox{90}{\textbf{MSE}}}& Cycle & 0.0314 $\pm$0.0006 & 0.5301$\pm$	0.4880  & 0.7072$\pm$	0.3956  & 0.1280$\pm$	0.1603 & 0.2437$\pm$	0.3111  & 0.0616$\pm$	0.0017 & 0.1887$\pm$	0.1565\\
 &CWRG& 0.7503	$\pm$0.1687 & 0.4607$\pm$	0.3602 & 0.6145$\pm$	0.4279 & 0.5803$\pm$	0.2688 & 0.6826$\pm$	0.2604& 0.5785$\pm$	0.2945 & 0.4825$\pm$	0.1722\\
 & SSA-CGAN-p & 0.0234 $\pm$0.0032& 0.0160 $\pm$0.0100& \textbf{0.0147$\pm$ 0.0018}& 0.0503$\pm$ 0.0051& 0.0262$\pm$  0.0017	& 0.0443 $\pm$0.0085& 0.0348$\pm$ 0.0021\\
 &SSA-CGAN & \textbf{0.0169$\pm$	0.0011} & \textbf{0.0102$\pm $	0.0024} &  0.0177$\pm$	0.0071 &  \textbf{0.0271$\pm$	0.0007} & \textbf{0.0202$\pm$	0.0014} &  \textbf{0.0210$\pm$	 0.0011}  & \textbf{0.0235$\pm$	0.0041}\\
\midrule
  \multirow{3}{*}{\rotatebox{90}{\textbf{MAE}}}& Cycle & 0.0608 $\pm$ 0.0041 &0.4924$\pm$	0.4146  & 0.6231$\pm$ 	0.3264  & 0.2162$\pm$ 0.1610 & 0.4236 	$\pm$ 0.2957  & 0.1409$\pm$ 0.0022 & 0.3048$\pm$ 0.1939\\
& CWRG& 0.6963$\pm$ 0.3738 & 0.4564$\pm$ 0.3868 & 0.5603 $\pm$0.5564 & 0.6819$\pm$0.1240 & 0.7008 $\pm$0.1478	& 0.5258$\pm$0.2860 & 0.5189$\pm$0.2800\\
 & SSA-CGAN-p & 0.0508$\pm$0.0037& 0.0411$\pm$0.0118& \textbf{0.0390$\pm$0.0028}& 
               0.1322$\pm$0.0059& 0.0834$\pm$0.0029& 0.1155$\pm$0.0118& 0.0837$\pm$0.0048\\
 &SSA-CGAN & 	\textbf{0.0436$\pm$ 0.0011} & \textbf{0.0338$\pm$ 0.0046} &   0.0426$\pm$ 0.0089 & \textbf{0.0947$\pm$  	0.0018} 	 & \textbf{0.0720 $\pm$0.0043} &  	 	\textbf{0.0754 	$\pm$ 0.0026}  & \textbf{0.0613 $\pm$0.0069}\\
\bottomrule
\end{tabular}
}
\end{adjustwidth}
\end{center}
\caption{MSE and MAE for various paired transformations averaged across five runs with one standard deviation.}
\label{tab:main_results}
\end{table}

\begin{figure}[t]
    \centering
    \includegraphics[width=\linewidth]{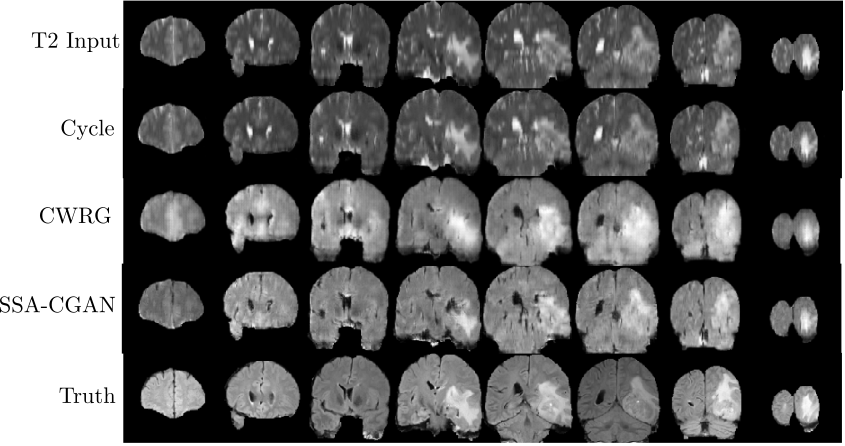}
   \caption{A comparison of the transformation from T2 to FLAIR.}
\label{fig:t2_flair_comparison}
\end{figure}
   \setlength{\belowcaptionskip}{-0.5cm} 
\begin{figure}[t]
    \centering
    \includegraphics[width=\linewidth]{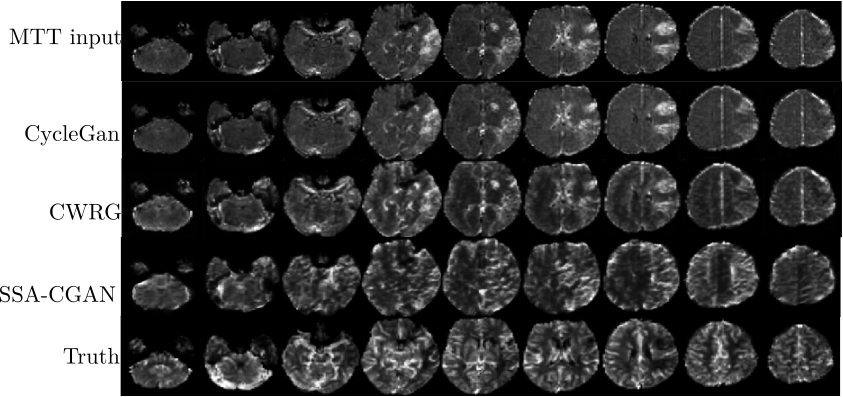}
    
   \caption{A comparison of the transformation from MTT to CBF.}

\label{fig:isles_MR_rCBF_MR_MTT_comparison}
\end{figure}
Results for the performance of \emph{SSA-CGAN} are shown in Table \ref{tab:main_results}. We observe that the \emph{SSA-CGAN} yields from a 8.32\% reduction from the CycleGAN (T2 to T1) up to a 89.6\% decrease in MSE in the case of CBF to CBV with an average reduction of 33.8\% and 46.0\% in MAE and MSE respectively across all transformations. The consistent out-performance of our method over the CycleGAN demonstrate there is potential gains when the information from paired data points can be leveraged. This is further emphasised by the improvement over SSA-CGAN-p which has been trained using only \emph{paired} data. By leveraging \emph{unpaired} data during training, the \emph{SSA-CGAN} produces a reduction of 18.02\% and 28.16\% in MAE and MSE on average when compared to SSA-CGAN-p. SSA-CGAN produces a lower MSE in most cases despite CWRG includes a loss component that minimises the $l_2$ norm. Furthermore, \emph{SSA-CGAN} produces lower variance compared to other methods demonstrating that our method is less sensitive to different weight intializations and improves the stability of training and convergence. 

Figure \ref{fig:t2_flair_comparison} and \ref{fig:isles_MR_rCBF_MR_MTT_comparison} shows a comparison of the transformation from T2 to FLAIR and MTT to CBF respectively, of a particular chosen MR scan produced by the various models. The CycleGAN produces no noticeable change from the input image and the CWRG creates a smoothed version of the ground truth. This can be attributed to the MSE component of the objective function where the MSE pushes the generator to produce blurry images\cite{mathieu2015deep}.  The additional adversarial component of our method forces the generator to synthesise a more visually realistic image. However, in Figure \ref{fig:t2_flair_comparison} the image produced does not match the pixel intensity of the ground truth and in Figure \ref{fig:isles_MR_rCBF_MR_MTT_comparison}, fails to capture the high detail and edges of the CBF modality and fails to distinguish between background and low intensity areas.

\setlength{\belowcaptionskip}{0cm}
\begin{figure*}
\begin{center}
\includegraphics[width=\linewidth]{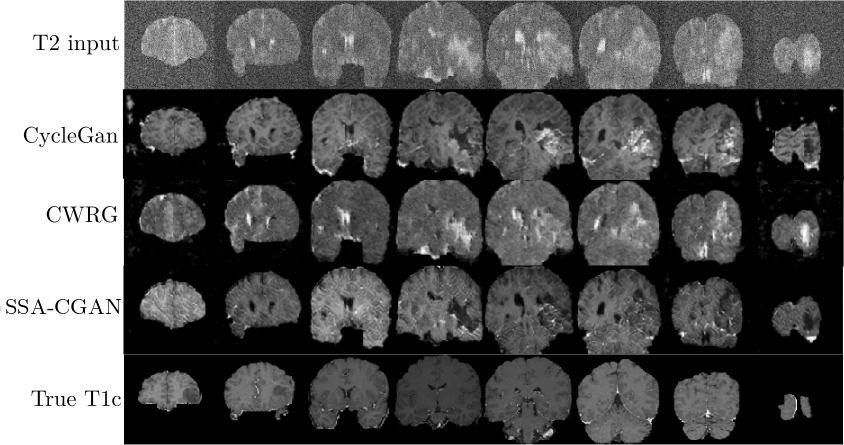}

\end{center}
   \caption{A T2 image was corrupted with Gaussian noise and was transformed to a T1c image by the various models.}
\label{fig:brats_t1c_t2_noise_comparison}
\end{figure*}

\subsection{Robustness to noise}

\begin{wrapfigure}[16]{l}{0.5\textwidth}

\centering
\includegraphics[width=\linewidth]{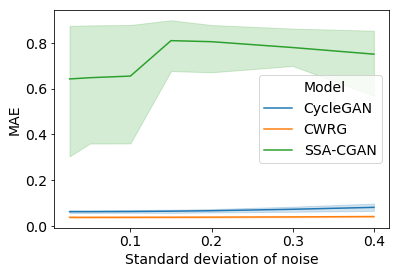}
   \caption{Quantitative comparison of the reconstruction error by varying the amount of random noise injected to test data.}
\label{fig:brats_t2_flair_noise_mae}
\end{wrapfigure}

The methods  were assessed by injecting random Gaussian noise into the test data to simulate thermal noise conditions to evaluate the robustness of the models, despite not being trained on noisy examples. Various levels of noise was injected to the data, ranging from a standard deviation of $0.025$ to $0.4$. The predictions of the models was evaluated against the ground truth. Figure \ref{fig:brats_t2_flair_noise_mae} shows the comparison between the models, with the MAE as the evaluation metric. At all noise levels, the \emph{SSA-CGAN} outperforms other methods with lower variance further demonstrating the robustness of our method.

The methods were also visually evaluated under extreme simulated thermal noise conditions by adding Gaussian noise with mean 0 standard deviation of 0.2 to the input. Figure \ref{fig:brats_t1c_t2_noise_comparison} shows the transformation produced by a noisy input volume to the networks. The CWRG produces noise filtered version of the T2 scan and fails to perform the transformation to T1c. Our method and the CycleGAN shows robustness under the extreme scenario and fabricates successful slices. However, it fails to hide the tumour in the T2 scan (the bright spot in bottom right) in the T1c reconstruction and instead substitutes background for that tumour.

\subsection{Limitations}
This approach has several limitations. Due to the additional discriminator that distinguishes paired examples, additional computational time is required for training. Second, adversarial networks remain a very active area of research, and are known to be difficult to train and suffer issues such as mode collapse\cite{goodfellow2016nips}. Further work would be to investigate the effect on performance when the fraction of paired examples changes and  the point where the paired-input discriminator fails to be effective.

\section{Conclusion}
Many state-of-the-art models in brain tissue segmentation and disease classification require multiple modalities during training and inference. However, examples where all modalities are available is limited and therefore the ability to incorporate unpaired data could be important for the adoption of these methods in clinical settings or improve existing models. Furthermore, the overall data avilable in limited and MRI volumes are high dimensional. The \emph{Semi-Supervised Adversarial CycleGAN} (SSA-CGAN) learns translations between neuroimaging modalities using \emph{unpaired} data and \emph{paired} examples through a \textit{cycle-consistency} loss, an adversarial signal for the discrimination between generated and real images of each domain and an additional adversarial signal that discriminates between the pairs of real data and pairs of generated images. Our experimental results have demonstrated that \emph{SSA-CGAN} has superior results in achieving lower reconstruction error and is more robust compared to all of current state-of-the-art approaches across a wide range of modality translations. 
%
%
%
%
%
%
\bibliographystyle{splncs04}
\bibliography{egbib}

\end{document}